\documentclass[runningheads]{llncs}
\usepackage[T1]{fontenc}
\usepackage{graphicx,verbatim}
\usepackage{amsmath}
\usepackage{amssymb}
\usepackage{textcomp}
\usepackage{url}
\begin{document}
\title{
Virtual 3D H\&E Staining from Phase-contrast Back-illumination Interference Tomography
}

%% Add paired in fluorescence and BIT in title bc valuable. 

\begin{comment}  %% Removed for anonymized MICCAI submission
\author{First Author\inst{1}\orcidID{0000-1111-2222-3333} \and
Second Author\inst{2,3}\orcidID{1111-2222-3333-4444} \and
Third Author\inst{3}\orcidID{2222--3333-4444-5555}}
%
\authorrunning{F. Author et al.}
\institute{Princeton University, Princeton NJ 08544, USA \and
Springer Heidelberg, Tiergartenstr. 17, 69121 Heidelberg, Germany
\email{lncs@springer.com}\\
\url{http://www.springer.com/gp/computer-science/lncs} \and
ABC Institute, Rupert-Karls-University Heidelberg, Heidelberg, Germany\\
\email{\{abc,lncs\}@uni-heidelberg.de}}

\end{comment}

\author{Anthony A. Song\inst{1} \and Boyan Zhou\inst{1} \and Mayank Golhar\inst{1} \and Marisa Morakis\inst{1} \and Alexander Baras\inst{2} \and Nicholas J. Durr\inst{1}} %\and \author{Mayank Golhar} \and \author{Marisa Morakis} %\and \author{Alexander Baras} \and \author{Nicholas J. Durr}   %% Added for anonymized MICCAI submission
%\and \author{Anthony A. Song\inst{1}\orcidID{0000-0002-0935-509X}}
\authorrunning{Anthony A. Song et al.}
\institute{Department of Biomedical Engineering, Johns Hopkins University, \\
    Baltimore, MD, USA\\
    \email{asong18@jhu.edu, ndurr@jhu.edu} \and 
    Department of Pathology, Johns Hopkins Hospital, \\
    Baltimore, MD, USA\\
    }
  
\maketitle              % typeset the header of the contribution
\begin{abstract}
Three-dimensional (3D) histopathology of unprocessed tissues has the potential to transform disease management by enabling volumetric characterization of tissue microarchitecture and in-vivo assessment. Back-illumination Interference Tomography (BIT) is a new phase microscopy technology that provides rapid, non-destructive volumetric imaging of unprocessed tissues. However, translating BIT volumes into clinically interpretable H\&E images remains challenging, particularly due to shift-variant contrast and the absence of quantitative validation benchmarks. We introduce \textbf{HistoBIT3D}, the first voxel-wise paired BIT and fluorescence-labeled nuclei dataset, enabling quantitative evaluation of structural preservation in unsupervised virtual staining against ground-truth nuclear distributions. Using this dataset, we present a novel virtual staining framework that translates BIT volumes with shift-variant contrast into realistic H\&E volumes by leveraging bidirectional multiscale content consistency and cross-domain style reuse to enhance structural fidelity and perceptual realism. Our method achieves state-of-the-art realism metrics while significantly improving 3D nuclei segmentation accuracy and boundary preservation under zero-shot Cellpose evaluation. Together, these contributions establish a quantitatively validated, structurally faithful, and scalable pipeline for 3D virtual H\&E staining, advancing the paradigm of slide-free, volumetric computational histopathology. Our data and code are available at: \\ \url{https://github.com/aasong113/HistoBIT3D\_VirtualStaining}.


% Authors must provide keywords and are not allowed to remove this Keyword section.

\end{abstract}

\section{Introduction}

Histopathology is the gold standard for identifying structural and cellular changes that differentiate normal from diseased tissue. The current workflow uses hematoxylin and eosin (H\&E) staining of sliced two-dimensional (2D) tissue samples and analysis under bright-field microscopy. However, 2D tissue samples only give a small glimpse of the complex three-dimensional (3D) morphological structures. A shift towards 3D pathology may allow for better characterization of tissue microstructures for downstream pathological understanding. 3D pathology has advanced the study of pancreatic tumorigenesis and cardiovascular microanatomy using serially sectioned H\&E reconstruction (CODA)\cite{Kiemen2022CODA,Kathiriya2026DisruptedBoundary}, enabled clinical risk assessment in prostate cancer and Barrett esophagus via open-top light-sheet microscopy\cite{Serafin2023NonDestructive3D,Reddi2023BarrettEsophagusAtlas}, and supported skin-biopsy analysis using two-photon fluorescence microscopy\cite{ChingRoa2022RealtimeTPFM}. However, current 3D histopathological techniques require complex instrumentation and labor-intensive sample preparation, including sectioning, staining, or optical clearing, making them difficult to deploy for real-time intraoperative analysis and incompatible with in-vivo imaging. To address these limitations, approaches based on intrinsic phase contrast, such as optical coherence tomography (OCT) \cite{Yin2025FastLabelFree3D}, oblique back-illumination microscopy (OBM) \cite{Ford2012PhaseGradient}, and back-illumination interference tomography (BIT) \cite{McKay2025BackIllumination}, have emerged as label-free 3D imaging modalities.

A major challenge with pathology applications of phase microscopy is that the contrast is fundamentally different from that of H\&E. Phase microscopy maps refractive index gradients, while H\&E captures the distribution of color stains mediated by chemical and molecular interactions. Generative adversarial networks (GANs) have been developed to generate "virtually stained" H\&E-like images from other microscopy techniques\cite{Abraham2023LabelFree3DHisto,Yin2025FastLabelFree3D,AlmagroPerez2025HistologyGuidedMicroCT,Song2025SlideFreeHisto}. Initially, virtual staining employed supervised learning strategies and was shown to successfully reconstruct histological features across the input and target domains\cite{Park2025Holotomography3D,Rivenson2019VirtualHistology}. However, this requires precisely aligned supervised image pairs from both imaging modalities, which is often impractical due to tissue degradation, staining quality, and hardware limitations. To overcome this limitation, CycleGAN \cite{Zhu2017CycleGAN} has become a popular unsupervised framework that allows virtual staining of thick, unstained tissue using corresponding unpaired H\&E images \cite{Zhang2022HighThroughputHistology,Abraham2023LabelFree3DHisto,Song2025SlideFreeHisto,Yin2025FastLabelFree3D}. Cycle consistency encourages preservation of spatial structure but often fails to achieve precise pixel-wise alignment during translation. Extensions to CycleGAN have addressed this limitation by incorporating saliency consistency constraints \cite{Li2021UnsupervisedContentPreserving} and feature map information consistency losses \cite{You2025Stable}, improving structural fidelity in virtual staining tasks. However, these approaches have primarily been validated on modalities exhibiting shift-invariant contrast, such as fluorescence and differential interference contrast. BIT images, however, have shift-variant contrast that changes with the specimen’s position relative to the microscope. Furthermore, there are currently no methods for quantitatively benchmarking the results of unsupervised 3D virtual H\&E staining. Qualitative metrics like Fr\'{e}chet Inception Distance (FID) and Kernel Inception Distance (KID) provide little insight into content preservation during image translation. 

To address these concerns, we introduce HistoBIT3D, the first voxel-wise registered 3D dataset of BIT and fluorescence-labeled nuclei, and an innovative GAN framework for realistic, content-preserving virtual H\&E staining of BIT volumes. By introducing a 3D quantitative validation dataset with a novel architecture focused on content preservation and realism, we establish a foundation for 3D, unsupervised, fast, simple, stain-free, high-fidelity pathology imaging. The three main contributions are summarized as follows: 

\textbf{(1) The first voxel-wise paired 3D Back-illumination Interference Tomography (BIT) and fluorescent nuclei dataset}. This dataset enables quantitative validation of structural preservation in 3D virtual staining against ground-truth nuclear distributions.

\textbf{(2)A novel virtual staining framework for translating BIT volumes with shift-variant contrast into virtual H\&E volumes.} Our model incorporates a bidirectional multiscale content consistency loss that aligns hierarchical features across translation cycles to enforce structural fidelity.

\textbf{(3) Enhanced perceptual realism through efficient cross-domain style reuse}. We leverage AdaIN-based injection of learned H\&E style features into the BIT-to-H\&E generator to improve visual fidelity.

%%%% FIGURE 1: Overview. 
\begin{figure}
\includegraphics[width=\textwidth]{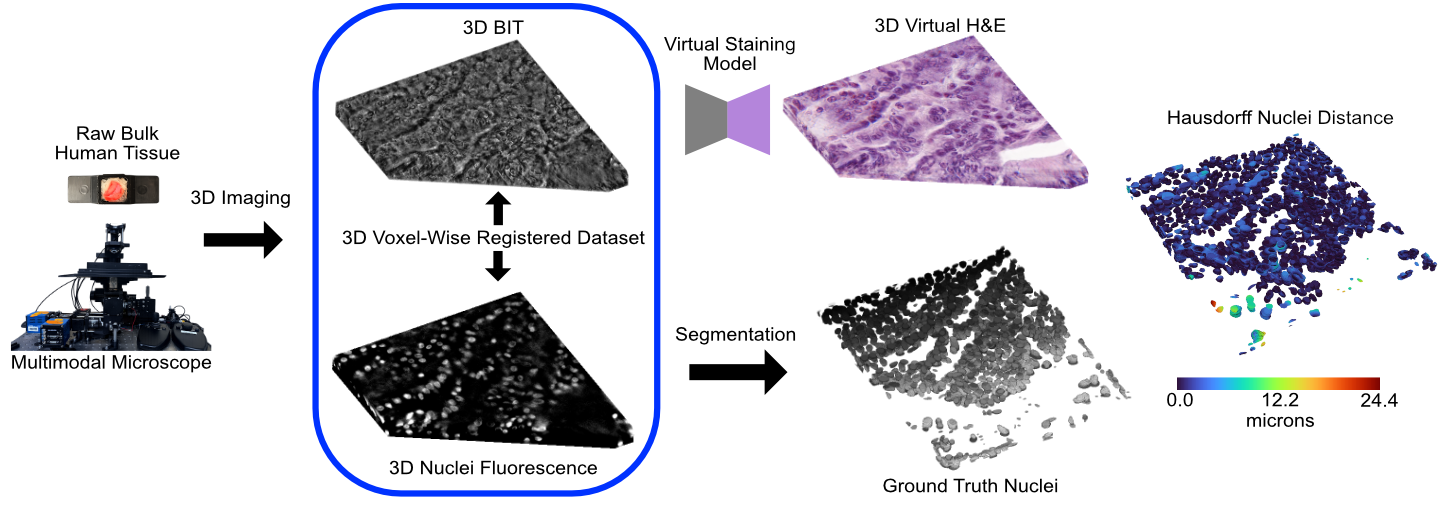}
\caption{Overview of the 3D virtual staining pipeline for Back-illumination Interference Tomography (BIT). From left to right: optical setup for volumetric acquisition of voxel-wise paired BIT and fluorescence nuclei data from bulk human tissue; virtual H\&E generation using our GAN-based framework (top row); 3D fluorescence nuclei segmentation results (bottom row); and quantitative evaluation via Hausdorff distance comparing segmented nuclei from virtual H\&E with ground-truth fluorescence volumes.}\label{fig1}
\end{figure}

\section{Methodology}

\subsection{Multi-modal BIT and Fluorescence 3D Pathology Dataset}

% We introduce HistoBIT3D, a multimodal dataset of slide-free tissue volumes that combines Back-illumination Interference Tomography (BIT) with voxel-wise paired 3D fluorescence imaging of nuclei and cytoplasmic structures, along with unpaired 2D FFPE H\&E images as the virtual staining target. To the best of our knowledge, this is the first dataset providing volumetric fluorescence ground truth registered to label-free imaging, enabling quantitative benchmarking of structural preservation in virtual staining methods. By providing objective validation of histological feature consistency—particularly nuclear morphology—HistoBIT3D supports the development and rigorous evaluation of unsupervised slide-free computational histopathology pipelines. The HistoBIT3D dataset includes voxel-wise paired BIT and nuclear fluorescence volumes of duodenum tissue with corresponding unpaired 2D H\&E images, additional 3D BIT-only duodenum datasets (crypts and submucosa) with matched 2D H\&E, and a 2D normal and cancer kidney dataset to evaluate cross-tissue generalization, with each subset containing approximately 5{,}000 images of size $512 \times 512$.

We introduce HistoBIT3D, a multimodal slide-free tissue dataset combining Back-illumination Interference Tomography (BIT) with voxel-wise paired 3D fluorescence imaging of nuclei and cytoplasmic structures, alongside unpaired 2D FFPE H\&E images for virtual staining. To our knowledge, this is the first dataset providing volumetric fluorescence ground truth registered to label-free imaging, enabling quantitative benchmarking of structural preservation, particularly nuclear morphology, in unsupervised virtual staining. HistoBIT3D includes paired BIT–nuclear fluorescence duodenum volumes with corresponding unpaired 2D H\&E, additional 3D BIT-only duodenum datasets (crypts and muscularis) with matched 2D H\&E, and a 2D normal and cancer kidney dataset for cross-tissue generalization, with each subset comprising approximately 5{,}000 images of size $512 \times 512$.

BIT imaging was performed using 660\,nm LED back-illumination through the objective back aperture to generate transmission-like phase contrast from interference between diffracted and backscattered light near the focal plane. Registered fluorescence images of bulk tissue stained with Hoechst (nuclei) and eosin (cytoplasm) were acquired through the same optical path and camera under oblique UV excitation \cite{Fereidouni2017MUSE}. Volumetric stacks spanning 30--50\,\textmu m were collected for both modalities. BIT images were background-subtracted ($\sigma=30$) and scaled to 8-bit for training. To reduce shift-variant contrast due to focal-plane dependence, the network input consisted of a three-channel stack (original, inverted, original) to emphasize structural over intensity information. Fluorescence volumes were processed with 3D PSF deconvolution to suppress bulk scattering.

%%%% FIGURE 2: Network Architecture. 
\begin{figure}
\includegraphics[width=\textwidth]{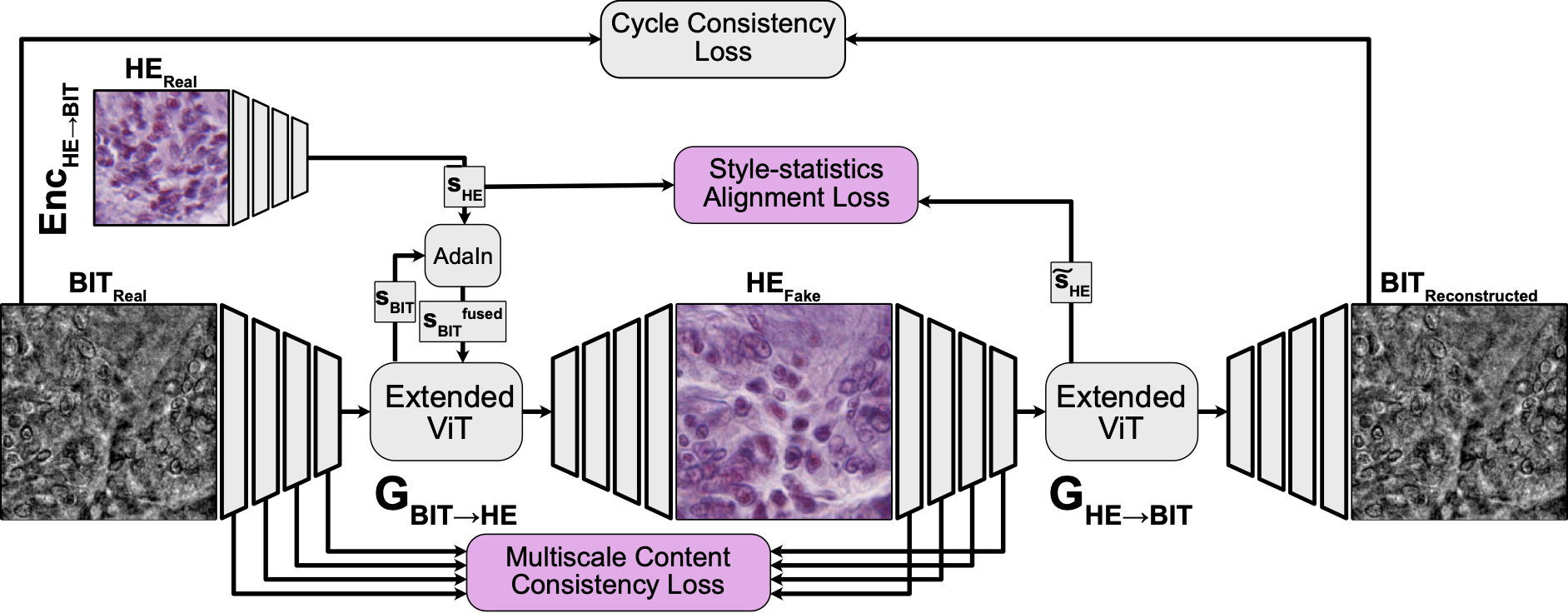}
\caption{%Architecture for multiscale content consistency and cross-domain style injection.}
Architecture showing bidirectional multiscale content consistency for structurally faithful BIT-to-H\&E translation and AdaIN-based cross-domain style injection for enhanced realism.}

\label{fig2}
\end{figure}

\subsection{Vision Transformer CycleGAN}

Prior work has demonstrated that CycleGAN-based architectures \cite{Abraham2023LabelFree3DHisto}, including variants with spatial consistency constraints \cite{Li2021UnsupervisedContentPreserving,You2025Stable}, are effective for virtual H\&E staining of label-free phase-contrast data due to their ability to preserve feature correspondence through cycle consistency while translating domain-specific appearance. However, biomedical imaging demands preservation of global tissue architecture beyond local texture realism, motivating the incorporation of non-local pattern learning mechanisms \cite{Dosovitskiy2020ImageIsWorth}. Vision transformers (ViTs) have shown improved modeling of long-range structural dependencies, such as relative nuclear arrangement and glandular organization, in virtual staining tasks \cite{Hussain2026ViTStain}. 
To capture global context, we build off UVCGANv2 \cite{Torbunov2023UVCGANv2}, which extends the classical CycleGAN framework with a ViT-based generator architecture. We enforce structural preservation using multiscale feature maps from the U-Net encoder together with spatial tokens from the ViT bottleneck as a content-consistency objective. Furthermore, UVCGANv2’s extended transformer design disentangles spatial and style representations, enabling source-driven style modulation through the reuse of target-domain style tokens to enhance staining realism while maintaining geometric fidelity. In this training paradigm, our model performs 2D-to-2D virtual staining of an axial stack, enabling 3D histopathology.

\subsection{Multiscale Content Consistency Loss}

To preserve content fidelity between virtually stained outputs and input BIT images, the learned feature representations must retain precise spatial information from the input domain \cite{You2025Stable}. In virtual staining, this is particularly important because the desired transformation is primarily a domain/style change, while underlying tissue morphology should remain unchanged. Motivated by hierarchical representation learning in modern vision backbones \cite{Chen2022ScalingVisionTransformers}, we introduce a bidirectional multiscale content consistency loss that aligns intermediate generator features across both translation cycles. An additional motivation is shift-variant contrast in BIT imaging: nuclei can appear dark or light depending on focal-plane position, even when their underlying morphology is unchanged.

% \begin{equation}
% \hat{x}_{\mathrm{BIT}} = G_{\mathrm{H\&E}\rightarrow \mathrm{BIT}}(G_{\mathrm{BIT}\rightarrow \mathrm{H\&E}}(x_{\mathrm{BIT}})), \hat{y}_{\mathrm{H\&E}} = G_{\mathrm{BIT}\rightarrow \mathrm{H\&E}}(G_{\mathrm{H\&E}\rightarrow \mathrm{BIT}}(y_{\mathrm{H\&E}}))
% \end{equation}

Our model consists of two generators \(G_{\mathrm{BIT}\rightarrow \mathrm{H\&E}}\) and 
\(G_{\mathrm{H\&E}\rightarrow \mathrm{BIT}}\) operating on inputs 
\(x_{\mathrm{BIT}} \in \mathbb{R}^{H \times W \times 3}\) and 
\(y_{\mathrm{H\&E}} \in \mathbb{R}^{H \times W \times 3}\). 
Let \(\phi_k^{\mathrm{BIT}\rightarrow\mathrm{H\&E}}(\cdot)\) and 
\(\phi_k^{\mathrm{H\&E}\rightarrow\mathrm{BIT}}(\cdot)\) denote the feature maps extracted from the \(k\)-th encoder stage of the respective generators, corresponding to downsampling scales \(k \in \{1,2,4,16\}\). The full-resolution map (\(k=1\)) preserves fine nuclear boundaries and micro-texture, intermediate scales encode mid-level morphology, and the \(\mathrm{ViT}\) bottleneck (\(k=16\)) captures global tissue organization, stabilizing histologic context despite focal-plane–dependent intensity shifts. %The full-resolution map (\(k=1\)) preserves fine spatial details (e.g., nuclear boundaries and micro-texture), intermediate scales encode mid-level morphology, and the \(\mathrm{ViT}\) bottleneck (\(k=16\)) captures global tissue organization. This hierarchy is particularly important for virtual staining, as shallow layers preserve nucleus-level geometry while deeper layers stabilize global histologic context despite focal-plane–dependent intensity shifts. 
To reduce memory overhead, we sample a subset of channels \(\mathcal{C}_s\) at later scales. We then enforce bidirectional feature consistency:
\begin{equation}
\mathcal{L}_{\mathrm{BIT}}
= \frac{1}{|\mathcal{K}|}\sum_{k\in\mathcal{K}}
\left\|
\phi_{k,\mathcal{C}_s}^{\mathrm{BIT,H\&E}}(x_{\mathrm{BIT}})
-
\operatorname{sg}\!\left(\phi_{k,\mathcal{C}_s}^{\mathrm{H\&E,BIT}}(G_{\mathrm{BIT}\rightarrow \mathrm{H\&E}}(x_{\mathrm{BIT}}))\right)
\right\|_1,
\end{equation}
\begin{equation}
\mathcal{L}_{\mathrm{H\&E}}
= \frac{1}{|\mathcal{K}|}\sum_{k\in\mathcal{K}}
\left\|
\phi_{k,\mathcal{C}_s}^{\mathrm{H\&E,BIT}}(y_{\mathrm{H\&E}})
-
\operatorname{sg}\!\left(\phi_{k,\mathcal{C}_s}^{\mathrm{BIT,H\&E}}(G_{\mathrm{H\&E}\rightarrow \mathrm{BIT}}(y_{\mathrm{H\&E}}))\right)
\right\|_1.
\end{equation}

The multiscale consistency term is 
\(\mathcal{L}_{\mathrm{msc}}=\mathcal{L}_{\mathrm{BIT}}+\mathcal{L}_{\mathrm{H\&E}}, \quad 
\mathcal{L}=\lambda_{\mathrm{msc}}\mathcal{L}_{\mathrm{msc}}\).
Here, \(\operatorname{sg}(\cdot)\) denotes the stop-gradient operator, which treats one branch as a fixed target while updating the opposite branch to improve optimization stability.
%%%%%%%%%%%%%%%%%%%%%%%%%%%%%%%%%%%%%%%%%%%%%%%%%%%%%%%%%%%%%%%%%%%%%%%%%%%%%%%%%%%%%%%%%%%%%%%

\subsection{Cross-Domain Style Fusion }

UVCGANv2 promotes explicit style–content disentanglement at the transformer bottleneck by separating global appearance information from spatially structured content features, enabling controlled style modulation via Adaptive Instance Normalization (AdaIN)\cite{HuangBelongie2017}. In this formulation, the injected bottleneck token is interpreted as a global style token rather than a spatial content token, allowing appearance statistics to be modulated independently of tissue geometry. 
During reverse passes \(G_{\mathrm{H\&E} \rightarrow \mathrm{BIT}}\) on real H\&E samples \(y_{\mathrm{H\&E}}\), we extract the final extra token \(s_{\mathrm{H\&E}}\), representing the current H\&E style embedding at the ViT bottleneck. We update a running-mean prototype \(\bar{s}_{\mathrm{H\&E}}\) via exponential moving average with factor \(\alpha\). This prototype provides a stable estimate of the global H\&E style distribution, is checkpointed during training, and reused at inference to ensure consistent cross-dataset staining.

For forward passes \(G_{\mathrm{BIT} \rightarrow \mathrm{H\&E}}\) with input \(x_{\mathrm{BIT}}\), we fuse the BIT style token \(s_{\mathrm{BIT}}\) with \(\bar{s}_{\mathrm{H\&E}}\) via Adaptive Instance Normalization (AdaIN) \cite{HuangBelongie2017} in token space:
\begin{equation}
\tilde{s}_{\mathrm{BIT}} = \operatorname{AdaIN}(s_{\mathrm{BIT}}, \bar{s}_{\mathrm{H\&E}})
= \sigma(\bar{s}_{\mathrm{H\&E}})\frac{s_{\mathrm{BIT}} - \mu(s_{\mathrm{BIT}})}{\sigma(s_{\mathrm{BIT}})} + \mu(\bar{s}_{\mathrm{H\&E}}),
\end{equation}
where \(\mu(\cdot)\) and \(\sigma(\cdot)\) compute statistics across the embedding dimension. 
To gradually reduce style injection during training, the fusion weight follows a cosine decay schedule.
This mechanism enables \(\mathrm{BIT} \rightarrow \mathrm{H\&E}\) synthesis to retain a stable \(\mathrm{H\&E}\)-domain style signature without requiring paired \(\mathrm{H\&E}\) images at inference time~\cite{Karras2019}.

To further align \(\mathrm{H\&E}\)-domain style distributions, we introduce a bottleneck style-statistics loss. Let \(s_{\mathrm{H\&E}}^{r}\) denote the style token from \(G_{\mathrm{H\&E} \rightarrow \mathrm{BIT}}(y_{\mathrm{H\&E}})\), cached with stop-gradient, and \(s_{\mathrm{H\&E}}^{f}\) the token from the synthesized pathway \(G_{\mathrm{H\&E} \rightarrow \mathrm{BIT}}(G_{\mathrm{BIT} \rightarrow \mathrm{H\&E}}(x_{\mathrm{BIT}}))\). We match first- and second-order statistics between real and generated style tokens via
\begin{equation}
\mathcal{L}_{\text{style}}
= \|\mu(s_{\mathrm{H\&E}}^{f}) - \operatorname{sg}(\mu(s_{\mathrm{H\&E}}^{r}))\|_2^2
+ \|\sigma(s_{\mathrm{H\&E}}^{f}) - \operatorname{sg}(\sigma(s_{\mathrm{H\&E}}^{r}))\|_2^2,
\end{equation}
This encourages the generated branch to match the real \(\mathrm{H\&E}\) style distribution while restricting gradient flow to the synthesized pathway~\cite{HuangBelongie2017}.

%%%%%%%%%%%%%%%%%%%%%%%%%%%%%%%%%%%%%%%%%%%%%%%%%%%%%%%%%%%%%%%%%%%%%%%%%%%%%%%%%%%%%%%%%

\section{Experiments and Results}

\subsection{Implementation Details}

We pretrained the generator backbone as a joint masked autoencoder on unpaired BIT and FFPE H\&E images for 50 epochs, with a batch size of 4. The full model was trained on BIT input and the FFPE-HE target style domain, using $60\%$ of the $512\times 512$ patches from our HistoBIT3D dataset with a batch size of 2 on a single NVIDIA GeForce RTX 3090 GPU for 170 epochs. Loss weights are $\lambda_{\text{cycle}}=10$, $\lambda_{idt}=0.5$, $\lambda_{msc}=1.0$ , and $\lambda_{style}=1.0$.

\subsection{Realism and 3D content preservation}

To validate realistic and content-preserving stain transfer, we compared our model with state-of-the-art virtual staining methods, including CycleGAN \cite{Zhu2017CycleGAN}, STABLE \cite{You2025Stable}, CycleDiffusion \cite{Wu2022CycleDiffusion}, and baseline UVCGANv2 (Base)\cite{Torbunov2023UVCGANv2}. We selected widely used unsupervised image translation metrics, FID and KID, to benchmark image quality and similarity. To assess the accuracy of virtual staining in 3D, we used zero-shot Cellpose \cite{Stringer2021Cellpose} to segment nuclei in our voxel-wise-registered fluorescent and virtual H\&E volumes ($130\mu m \times 83\mu m \times 20\mu m$). The fluorescent labeled nuclei volume acts as our ground truth nuclei locations, as Hoechst\cite{Fereidouni2017MUSE} is a well-known nucleic acid stain. We characterized the fidelity of our virtually stained 3D volume with conventional 3D segmentation metrics- foreground voxel-level 3D DICE, population-level 95th percentile Hausdorff Distance (HD95) on global foreground boundaries, and mean per-instance nuclei volume as a pathology-based morphological metric. We calculate 3D metrics by first translating our 2D masks into a 3D volume using u-Segment3D\cite{Zhou2025UniversalConsensus3D}.

%%%% FIGURE 3: Montage 
\begin{figure}
\includegraphics[width=\textwidth]{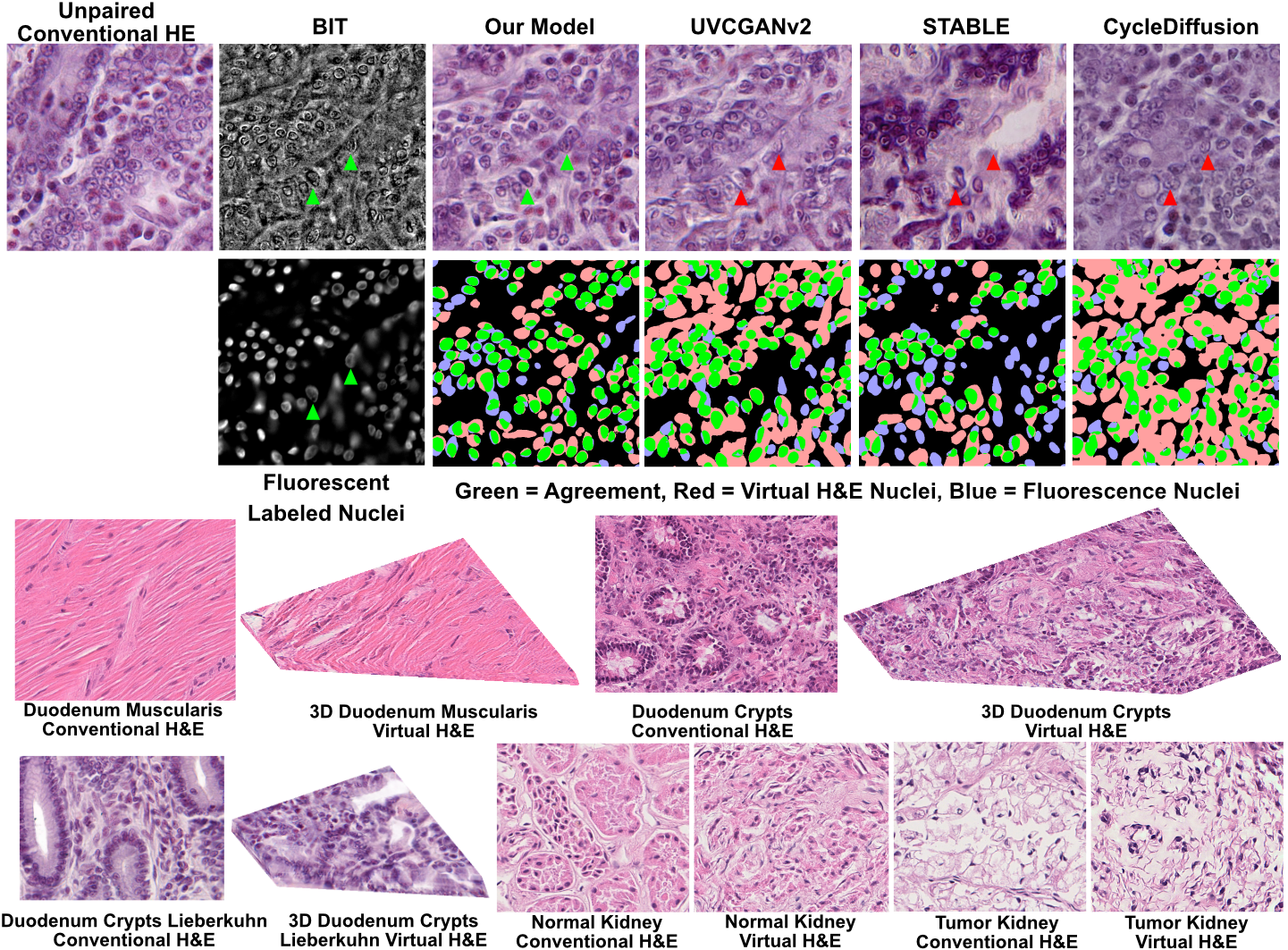}
\caption{\textbf{Row 1:} Comparison of virtual H\&E results from our method and baseline models against unpaired conventional H\&E. \textbf{Row 2:} Zero-shot Cellpose \cite{Stringer2021Cellpose} segmentation of virtual H\&E images compared with ground-truth nuclei from fluorescence imaging. Green arrows highlight nuclei whose structural content is preserved. \textbf{Rows 3–4:} Diverse tissue samples from the HistoBIT3D dataset, virtually stained from BIT into H\&E.}\label{fig3}
\end{figure}

The performance of our framework is demonstrated qualitatively in Fig.\ref{fig3} and quantitatively summarized in Table\ref{table1}. Row 1 of Fig.\ref{fig3} qualitatively compares our model with baseline approaches and conventional unpaired H\&E, which can be quantitatively validated by our model's FID (60.69) and KID (0.0417) outperforming baseline methods. Our model produces faithful translations of the structural features observed in the input BIT image, as confirmed by alignment with the corresponding fluorescent nuclei ground truth. Row 2 of Fig.\ref{fig3} demonstrates improved zero-shot Cellpose segmentation for our model, reflected by higher 3D Dice (0.594), nuclei volume consistent with ground truth fluorescence nuclei volume(408.3$\mu m^{3}$ vs. 405.6$\mu m^{3}$), and lower HD95 (4.04~$\mu m$), indicating superior boundary preservation. The success of zero-shot segmentation further suggests that the generated H\&E images are sufficiently realistic to be reliably recognized by Cellpose. Rows 3–4 of Fig.~\ref{fig3} present multiple virtually stained samples from the HistoBIT3D dataset, including duodenum tissue, normal kidney, and tumor kidney.

To evaluate the impact of cross-domain style fusion and multiscale content consistency, we conducted an ablation study starting from the UVCGANv2 base model and incrementally adding components: style injection and, finally, the full multiscale consistency loss (Our Model). We observe a substantial improvement in realism and 3D segmentation metrics with the addition of cross-domain style fusion, indicating higher-fidelity translation compared to the baseline, with further gains when combined with the multiscale content consistency loss.

\begin{table}[t]
\centering
\caption{Comparison with SOTA and ablation study for realism and fidelity.}
\label{table1}
\begin{tabular}{l|c|c|c|c|c|c}
  & FID$\downarrow$ & KID$\downarrow$ & 3D DICE & Nuc. Volume ($\mu m ^{3}$) & HD95 ($\mu m$) \\
\hline
CycleGAN      & 136.01 & 0.1556 & 0.360 & 377.2 &  5.59 \\
\hline
STABLE\cite{You2025Stable} & 86.54 & 0.0684 & 0.418 & 331.9  &  5.38 \\
\hline
CycleDiffusion\cite{Wu2022CycleDiffusion} & 106.83 & 0.1138 &  0.359&  233.2 & 4.77 \\
\hline
UVCGANv2\cite{Torbunov2023UVCGANv2} & 95.80 & 0.0821 & 0.515 & 386.7 & 4.67\\
\hline
Base+Style & \underline{62.45} & \underline{0.0434} & \underline{0.583} & \underline{416.6}& \underline{4.22}\\
\hline
\textbf{Our Model}  & \textbf{60.69} & \textbf{0.0417} & \textbf{0.594} & \textbf{408.3} & \textbf{4.04} \\
\hline
\end{tabular}

\textbf{*Bold} and \underline{underlined} indicate best and second best results, respectively. \\
Ground Truth Nuclei Volume is \textbf{405.6}$\mu m ^{3}$.
\end{table}

\section{Conclusions}

In this work, we advance 3D virtual histopathology by introducing HistoBIT3D, the first voxel-wise paired 3D BIT–fluorescence dataset, enabling quantitative validation of structural preservation against ground-truth nuclear distributions. We further propose a virtual staining framework that enforces structural integrity through bidirectional multiscale feature alignment and enhances perceptual realism via cross-domain style fusion. Future work will address contrast discrepancies between fluorescence and BIT to improve feature correspondence, incorporate fluorescence-guided semi-supervision, and refine downstream segmentation with task-specific tuning. To foster continued progress, HistoBIT3D will be made publicly available. Together, these contributions establish a validated and structurally faithful pipeline for 3D virtual H\&E staining, promoting the adoption of slide-free, volumetric computational histopathology.

 %% removed for anonymized MICCAI submission.
    
    % The following acknowledgement and disclaimer sections can be removed for the double-blind review process.  If and when your paper is accepted, reinsert the acknowledgement and the disclaimer clause in your final camera-ready version.
    % IF you opted to include the acknowledgement and disclaimer sections, they will count towards the 8-page limit.

% Embedded bibliography for arXiv submission to avoid undefined citations.

\end{document}